\documentclass[10pt]{article} 
\usepackage[preprint]{rlc}

\usepackage{amssymb}            
\usepackage{mathtools}          
\usepackage{mathrsfs}           
\usepackage{graphicx}           
\usepackage{subcaption}         
\usepackage[space]{grffile}     
\usepackage{url}                

\renewcommand{\S} {\mathcal{S}}
\newcommand{\A} {\mathcal{A}}

\newcommand{\M} {\mathcal{M}}

\usepackage{mdframed}
\newmdtheoremenv{theorem}{Theorem}

\newcommand{\qed}{\hfill \ensuremath{\blacksquare}}

\title{An Optimal Tightness Bound for the Simulation Lemma}

\author{Sam Lobel  \\
    samuel\_lobel@brown.edu \\
    Department of Computer Science \\
    Brown University
    \And
    Ronald Parr \\
    parr@cs.duke.edu\\
    Department of Computer Science \\
    Duke University
    }

\begin{document}

\maketitle

\begin{abstract}
We present a bound for value-prediction error with respect to model misspecification that is tight, including constant factors.
This is a direct improvement of the ``simulation lemma,’’ a foundational result in reinforcement learning.
We demonstrate that existing bounds are quite loose,
becoming vacuous for large discount factors,
due to the suboptimal treatment of compounding probability errors.
By carefully considering this quantity on its own, instead of as a subcomponent of value error, we derive a bound that is sub-linear with respect to transition function misspecification. We then demonstrate broader applicability of this technique, improving a similar bound in the related subfield of hierarchical abstraction.
\end{abstract}

\section{Introduction}
\label{sec:introduction}

In reinforcement learning, an agent is frequently tasked with making decisions in an environment that it cannot model perfectly. This may occur because the environment is learned about through sampled data, or because the agent's environment model is simplified through some abstraction. In such cases it is natural to ask, how might the quality of this approximation impact an agent’s decision making? This is the subject of the ``simulation lemma,’’ a foundational result in reinforcement learning that bounds the error in value estimation when the transition and reward function are known only with some specified degree of precision.

The simulation lemma was introduced in the context of exploration and finds use in a variety of domains that utilize imperfect models, such as hierarchical abstraction \citep{abel2016near} and offline policy evaluation \citep{yin2021near}. Frequently, results of this kind rely on developing a recursive relationship between the value error at subsequent timesteps. We show that this approach implicitly  overestimates how probability errors compound over time. By more directly approximating this quantity, we produce a bound on value-estimation error that is demonstrably tight. We then show that existing bounds can be derived as a linearization of our result, and finally apply our result to a hierarchical setting to demonstrate broader applicability. 

\section{Background and Related Work}

We develop our results in the framework of Markov Decision Processes (MDPs): $\M=(\S,\A,R, T,\gamma)$, where $\S$ is the state space, $\A$ is the action space, and $\gamma \in [0,1]$ is the discount factor. The next-state transition probabilities are given by $T(s'|s,a)$, and the reward function by $R(s,a) \in [0, 1]$.
A policy $\pi(a|s)$ gives the probability of taking an action from a given state. The objective in the MDP framework is generally either to construct a policy $\pi$ that maximizes the expected $\gamma$-discounted sum of reward, or to evaluate a given policy on this same measure.

When a model of the environment is given, these quantities can be computed exactly, for example through policy iteration or dynamic programming \citep{howard1960dynamic}. 
In reinforcement learning, however, the agent generally is not given this model, and instead must learn about the environment through interaction.
A common approach to this is model-based reinforcement learning \citep{moerland2023model,auer2006logarithmic}, which aims to estimate the environment's transitions and rewards from gathered data. However, when using finite data, the learned model is generally imperfect.
This work concerns itself with developing optimal bounds on policy evaluation error in the setting of misspecified models.
Here we detail a variety of areas in which such a bound is useful, along with related lines of study.

\paragraph{Exploration}
The original simulation lemma was introduced in the context of efficient exploration \citep{kearns2002near}, to quantify policy evaluation error as a function of state-action visitation counts. Understanding the effect of imperfect modelling is central to efficient exploration \citep{auer2006logarithmic,auer2008near,brafman2002r}. Methods that use these measures include count-based exploration \citep{strehl2008analysis} and its pseudocount approximations \citep{bellemare2016unifying,lobel2023flipping}.

\paragraph{Abstraction}
Model approximation frequently appears in the field of abstraction, where a full model of an MDP is replaced by one that is simpler in some respect. As we show later, our methodology can be used to improve the value error bounds when performing this replacement with state-action abstracted \textit{options} \citep{sutton1999between}.
A simple form of state abstraction is \textit{discretization}, where sets of states are grouped by some measure of similarity. A common example of this occurs in the \textit{partially observable} MDP framework \citep{lee2007makes,grover2021adaptive}, where the continuous belief-state space can be discretized into an approximate, finite MDP.

\paragraph{Offline Policy Evaluation}
The goal of offline policy evaluation (OPE) is to estimate the value of a policy using a fixed dataset of transitions, often generated by a different policy. Model-based OPE involves fitting an empirical model of transitions and rewards from this dataset, and using this to estimate value \citep{gottesman2019combining}. In this setting, the simulation lemma often is a key step in constructing accuracy bounds of the estimated value \citep{yin2020asymptotically,yin2021near}.

We also note that a variety of results in the literature bound the value error using different measures of similarity than the original simulation lemma.
Perhaps most closely related to our contribution is work that bounds multi-step transition error of imperfectly-modelled Lipschitz transition functions \citep{asadi2018lipschitz}. This results in a similar sum of compounding errors to ours, albiet in a different setting. Bisimulation metrics \citep{ferns2012metrics} unify transition and reward error into a single quantity that can be used to measure the similarity of MDPs with entirely different state spaces.

\section{Main Result}
\label{sec:method}

We begin by stating the conditions of the original simulation lemma. We consider two MDPs: ${\M=(\S,\A,R,T,\gamma)}$, and ${\hat{\M}=(\S,\A,\hat{R},\hat{T},\gamma)}$, which share a state-action space, but have (boundedly) different transition and reward functions. We are interested in the effect of running the same policy $\pi$ on these two related MDPs. Let $P^\pi$ be a matrix that contains the policy-conditioned state-state transition probabilities, and $R^\pi$ be a vector that contains the per-state expected reward:

\begin{equation}
\begin{aligned}
   P^\pi_{s,s'} \quad&=\quad \mathbb{E}_{a \sim \pi(s)} [T(s'|s,a)]\quad&=\quad \sum_{a \in \A}T(s'|s, a) \pi(a|s) \\
   R^\pi_s \quad&=\quad \mathbb{E}_{a \sim \pi(s)} [R(s,a)] \quad&=\quad \sum_{a \in \A}R(s,a)\pi(a| s).\quad
   \label{eq:p-pi-r-pi-definitions}    
\end{aligned}
\end{equation}

We define $\hat{P}^\pi$ and $\hat{R}^\pi$ analogously for MDP $\hat{\mathcal{M}}$. Throughout this work, a single index on a matrix  (or vector) extracts the specified row vector (or scalar). Furthermore, $P^a$ and $R^a$ refer to the transition probabilities, and expected reward, of executing action $a$ from each state. Using this notation, we can quantify the difference between two transition or reward functions with the following:

\begin{align}
\forall s, \pi: \; \lVert P^\pi_s - \hat{P}^\pi_s\rVert_1 \;\,\;\;&\le\;\; \epsilon_T \label{eq:original-sim-lemma-t-condition}\\
\forall \pi: \;\lVert R^\pi - \hat{R}^\pi\rVert_\infty \;\;&\le\;\; \epsilon_R.\label{eq:original-sim-lemma-r-condition}
\end{align}

We are interested in the value difference between running $\pi$ on each MDP. The value of a state for a given policy and MDP is defined as the expected discounted sum of rewards:
$$v^\pi(s) = \mathbb{E}_{a_i \sim \pi(s_i)}\big[\sum_{t=0}^\infty \gamma^t R(s_t, a_t)\; |\; s_0 = s, \M \big],$$

where $s_t$ is a random variable representing the state at timestep $t$.
Noting that ${\Pr(s_t=s'| s_0=s, \pi) = (P^\pi)^t_{s, s'}}$, we can concisely represent value in vectorized notation as follows:

$$V^\pi = \sum_{t=0}^\infty \gamma^t (P^\pi)^t R^\pi \quad,\quad V^\pi_s = \sum_{t=0}^\infty \gamma^t\langle (P^\pi)^t_s, R^\pi\rangle,$$

where $\langle \,\cdot\,, \,\cdot\, \rangle$ denotes the inner product between two vectors. We define $\hat{V}^\pi$ analogously for $\hat{\M}$.

\subsection{Original Simulation Lemma}

We are interested in quantifying the maximum value difference between running the same policy on two different MDPs. The original simulation lemma bounds this quantity as follows: 

\begin{equation}
    \forall s,\pi:\;\; |V^\pi_s - \hat{V}^\pi_s | \le \frac{\epsilon_R}{1 - \gamma} + \frac{\gamma\epsilon_T}{2(1 - \gamma)^2}.
    \label{eq:original-simulation-lemma-bound}
\end{equation}

Existing proofs of the simulation lemma frequently take advantage of a recursive representation of value (the Bellman Equation) \citep{howard1960dynamic}:
$$V^\pi = R^\pi + \gamma P^\pi \sum_{t=0}^\infty \gamma^t(P^\pi)^tR^\pi = R^\pi + \gamma P^\pi V^\pi.$$

For a complete proof, please refer to \citet{jiang2018notes} or see Appendix \ref{sec:appendix-full-old-sim-lemma-proof}. The key mathematical idea is to establish the following recursive relationship:
\begin{equation}
    \forall s,\pi:\;\;|V^\pi_s - \hat{V}^\pi_s| \le \epsilon_R + \frac{\gamma\epsilon_T}{2(1 - \gamma)} + \gamma \lVert V^\pi - \hat{V}^\pi\rVert_\infty,
    \label{eq:original-sim-lemma-recursion}
\end{equation}

which can then be easily transformed into the simulation lemma's bound.
Analyzing the recursive relationship above, the first term ($\epsilon_R$) represents a one-step reward-prediction error. The second term ($\frac{\gamma\epsilon_T}{2(1 - \gamma)}$) represents the maximum value error that results from misspecifying $\epsilon_T$ of the next-state distribution's probability mass.
However, by defining the recursive relationship as such, this bound implicitly assumes that the process can continually misspecify $\epsilon_T$ of its probability at each timestep. This quickly amounts to misspecifying more than the entire probability mass, leading to a vast overestimate of the value error, in particular when ${\epsilon_T > 1 - \gamma}$. In contrast, we carefully track the probability drift at each timestep to avoid this issue.

\subsection{Bounding Probability Distance}
We seek to bound the probability distance tightly at any timestep $t$. To do so effectively, it is useful to frame distances between probability vectors in terms of their overlap, instead of their $L_1$ distance.
We note that \cite{jiang2016structural} uses similar machinery to bound compounding probability error (Lemma 1), though applies this insight in a different context.
For two probability vectors $p,\,\hat{p}$, we define their overlap as $\bar{p}$, such that for each index $i$:
$$\bar{p}_i = \min(p_i, \hat{p}_i).$$
Usefully, because each element of $p - \bar{p}$ (and likewise $\hat{p} - \bar{p}$) is non-negative, the $L_1$ norm of the difference between these two vectors is equal to the difference between the $L_1$ norms:
\begin{equation}
\lVert p - \bar{p}\rVert_1 = \sum_i |p_i - \bar{p}_i| = \sum_i p_i - \sum_i\bar{p}_i = \lVert p \rVert_1 - \lVert\bar{p}\rVert_1
\label{eq:p-bar-identity}
\end{equation}
We use this to derive an equivalence between overlap and $L_1$ distance, related to the concept of \textit{total variation distance} \citep{levin2017markov}. Below, we use the notation $[p]^+$ to indicate a thresholded version of $p$ that retains only the non-negative parts, $[p]_i^+ = \max(p_i,0)$:
\begin{align}
    \lVert p - \hat{p}\rVert_1 \quad&= 
    \quad \lVert [p - \bar{p}]^+ \rVert_1 + \lVert [\hat{p} - \bar{p}]^+ \rVert_1 \nonumber\\
   &=\quad \lVert p - \bar{p} \rVert_1 + \lVert \hat{p} - \bar{p}\rVert_1\nonumber\\
    &=\quad \lVert p\rVert_1 - \lVert \bar{p}\rVert_1 + \lVert\hat{p}\rVert_1 - \lVert \bar{p}\rVert_1 \nonumber\\
    &=\quad 1 + 1 - 2\lVert \bar{p} \rVert_1 \nonumber\\
    \implies \lVert \bar{p} \rVert_1 \quad&=\quad 1 - \frac{\lVert p - \hat{p} \rVert_1}{2}.
    \label{eq:tvd-l1-equivalence}
\end{align}

\begin{figure*}[t!]
    \centering
    \includegraphics[width=0.6\linewidth]{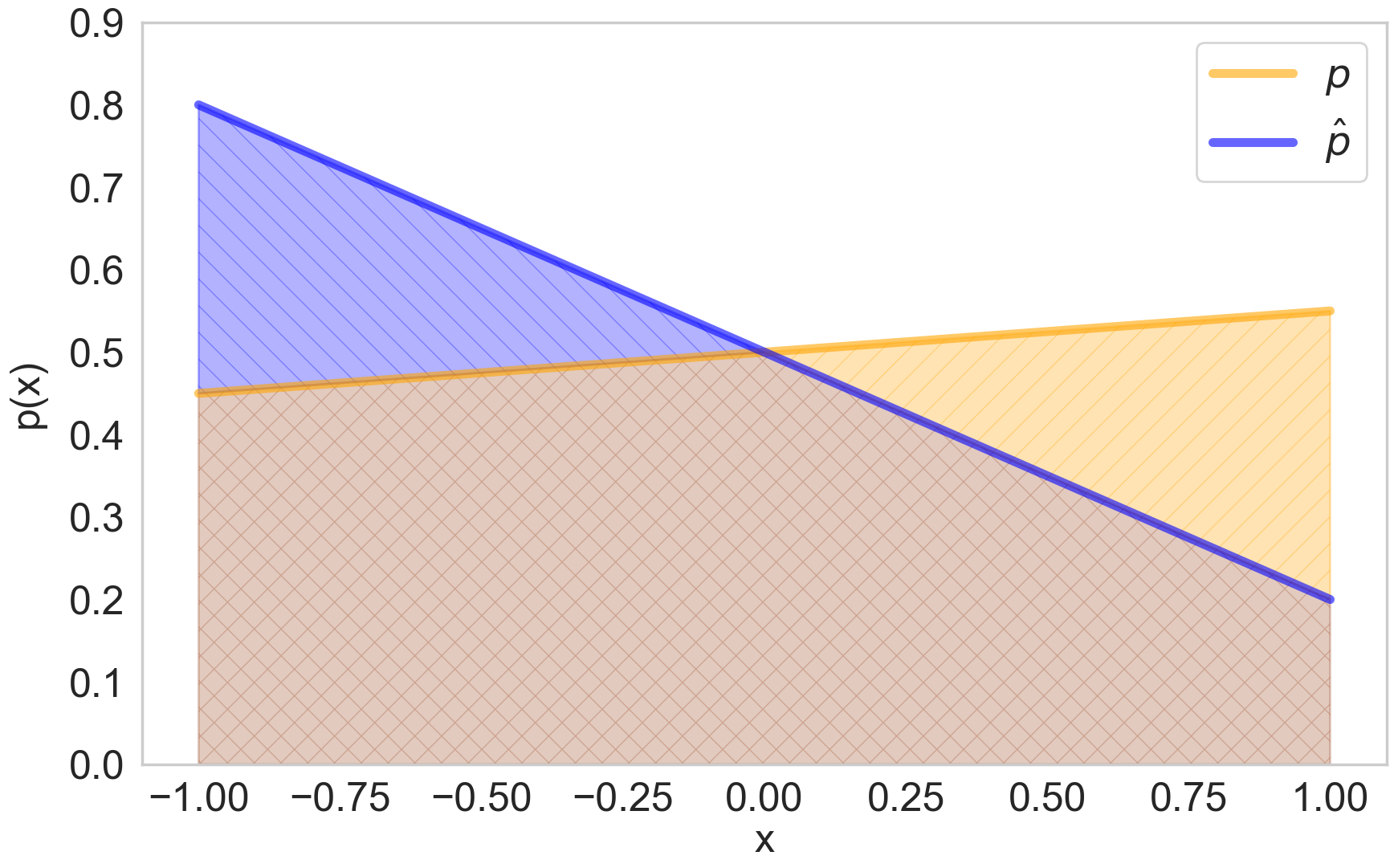}
    \caption{Visualization of relation between $L_1$ distance and overlap of two probability distributions (Equation \ref{eq:tvd-l1-equivalence}). The blue and orange shaded regions together comprise the $L_1$ distance. The brown region represents overlap. Overlap plus \textit{either} the blue or orange sections constitutes a probability distribution, and therefore has total area $1$. Thus the blue and orange regions both individually have area ${\lVert p - \hat{p}\rVert_1 / 2}$, and so ${\lVert \bar{p}\rVert_1 = 1 - \lVert p - \hat{p}\rVert_1 /2}$.}
    \label{fig:overlapping-probabilities}
\end{figure*}

See Figure \ref{fig:overlapping-probabilities} for a demonstration and explanation of this equivalence.
This relationship allows for a simple rewriting of the transition-error condition of the simulation lemma (Equation \ref{eq:original-sim-lemma-t-condition}):
\begin{equation}
    \forall s,\pi: \; \lVert\bar{P}^\pi_s\rVert_1 \ge 1 - \frac{\epsilon_T}{2}.\label{eq:tvd-sim-lemma-conditions}
\end{equation}

Using this framing, we can now lower-bound the overlap of state-distributions at timestep $t$ when starting from $s_0$, by demonstrating that at every timestep, at least $1 - \epsilon_T/2$ fraction of the prior timestep's distributional overlap is retained.
For notational convenience, ${P^t_{s_0,s} = (P^\pi)^t_{s_0,s}}$, and ${\bar{M}^t_{s_0,s} = \min(P^t_{s_0,s}, \hat{P}^t_{s_0,s})}$. Thus, 
\begin{align*}
    \lVert \bar{M}^{t+1}_{s_0} \rVert_1 \quad=&\quad\sum_{s'} \min(P^{t+1}_{s_0,s'}, \hat{P}^{t+1}_{s_0,s'}) \\
    =&\quad\sum_{s'} \min(\sum_{s}P^t_{s_0,s} \cdot P^\pi_{s,s'} \;, \; \sum_{s}\hat{P}^t_{s_0,s} \cdot \hat{P}^\pi_{s,s'}) \\
    \ge &\quad\sum_{s'}\sum_{s} \min(P^t_{s_0, s} \cdot P^\pi_{s,s'}\;, \; \hat{P}^t_{s_0, s} \cdot \hat{P}^\pi_{s,s'}) \\
    \ge &\quad\sum_{s'}\sum_{s} \min \Bigl( \min(P^t_{s_0,s}\;,\;\hat{P}^t_{s_0,s}) \cdot P^\pi_{s,s'})\;, \; \min(P^t_{s_0,s}\;,\;\hat{P}^t_{s_0,s}) 
    \cdot \hat{P}^\pi_{s,s'} \Bigr) \\
    = &\quad\sum_{s} \sum_{s'}\min(P^t_{s_0,s}\;,\;\hat{P}^t_{s_0,s})  \min(P^\pi_{s,s'}\;,\; \hat{P}^\pi_{s,s'}) \\
    = &\quad\sum_{s} \min(P^t_{s_0,s}\;,\;\hat{P}^t_{s_0,s}) \sum_{s'} \min(P^\pi_{s,s'}\;,\; \hat{P}^\pi_{s,s'}) \\
    \ge &\quad\lVert \bar{M}^t_{s_0} \rVert_1 \cdot \max_{s} \lVert \bar{P}^\pi_s \rVert_1 \\
    \implies \lVert \bar{M}^{t+1}_{s_0} \rVert_1 \quad\ge &\quad\lVert \bar{M}^t_{s_0} \rVert_1 \cdot (1-\epsilon_T/2).
\end{align*}

The third line can be understood as providing the minimum operator more options to choose from, in that after bringing the minimum inside of the sum, the two elements in the second line are both still possible choices and so the inequality holds. The fourth line can be understood similarly for multiplication.

With $\bar{M}^0 = I$ as the base case, applying recursion yields
\begin{equation}
    \lVert \bar{M}^t_{s_0}\rVert_1 \ge (1 - \epsilon_T/2)^t. \label{eq:recursive-magnitude-m}
\end{equation}

We contrast this with the equivalent recursive proof of distributional drift using the $L_1$ formulation of transition misspecification, akin to the recursion employed by the original simulation lemma (Equation \ref{eq:original-sim-lemma-recursion}):
\begin{align*}
    \lVert P^{t+1}_{s_0} - \hat{P}^{t+1}_{s_0} \rVert_1 \quad&=\quad \lVert P^{t}_{s_0} P^\pi - \hat{P}^{t}_{s_0} \hat{P}^\pi \rVert _1 \\
    &=\quad \frac{1}{2} \lVert (P^{t}_{s_0} - \hat{P}^{t}_{s_0})(P^\pi + \hat{P}^\pi) + (P^{t}_{s_0} + \hat{P}^{t}_{s_0})(P^\pi - \hat{P}^\pi)\rVert_1 \\
    &\le\quad \frac{1}{2} \lVert P^{t}_{s_0} - \hat{P}^{t}_{s_0} \rVert_1 \lVert (P^\pi + \hat{P}^\pi)^T\rVert_1 + \frac{1}{2}\lVert P^{t}_{s_0} + \hat{P}^{t}_{s_0} \rVert_1 \lVert (P^\pi - \hat{P}^\pi)^T\rVert_1 \\
    &=\quad \lVert P^{t}_{s_0} - \hat{P}^{t}_{s_0} \rVert_1 + \lVert P^\pi - \hat{P}^\pi\rVert_1 \\
    &\le\quad \lVert P^{t}_{s_0} - \hat{P}^{t}_{s_0} \rVert_1 + \epsilon_T \\
    \implies \lVert P^{t+1}_{s_0} - \hat{P}^{t+1}_{s_0} \rVert_1 \quad&\le\quad (t+1) \,\epsilon_T,
\end{align*}

where $\lVert \,\cdot\, \rVert_1$ above refers to both the matrix and vector $1$-norm, and on the third line we use the identity ${\lVert Ax \rVert_1 \le \lVert A \rVert_1 \lVert x \rVert_1}$.
This result makes clear the contrast between the two methods for computing distributional drift: Na\"ively using the $L_1$ formulation leads to unbounded accumulation of drift as horizon approaches infinity,
while the overlap formulation smoothly decays from $1$ to $0$.
This difference is crucial to generating the tighter bound in the next section.

\subsection{A Tight Bound on Value Error}
\label{subsec:main-proof}
We are now ready to prove our main result, a tight bound on the value error.
\newline
\begin{theorem}
    For two MDPs $\M$ and $\hat{\M}$ related as described in Equations \ref{eq:original-sim-lemma-t-condition} and \ref{eq:original-sim-lemma-r-condition}, the following inequality holds:
    \begin{equation}
        \forall s,\pi:\;\;|V^\pi_{s} - \hat{V}^\pi_{s}| \;\le\; \frac{1}{1 - \gamma} - \frac{1 - \epsilon_R}{1 - \gamma(1 - \epsilon_T/2)}.\label{eq:our-sim-lemma-bound}
    \end{equation}
    Furthermore, this bound is tight.
\end{theorem}

\textbf{Proof: } Since the conditions of the simulation lemma (Equations \ref{eq:original-sim-lemma-t-condition},\ref{eq:original-sim-lemma-r-condition}) are symmetric with respect to $\mathcal{M}$ and $\hat{\mathcal{M}}$, without loss of generality we assume $V^\pi_{s_0} \ge \hat{V}^\pi_{s_0}$, and thus $|V^\pi_{s_0} - \hat{V}^\pi_{s_0}| = V^\pi_{s_0} - \hat{V}^\pi_{s_0}$. We now add and subtract the same quantity in a way that allows for discarding a strictly non-positive term:

\begin{align*}
    |V^\pi_{s_0} - \hat{V}^\pi_{s_0}| \quad&=\quad V^\pi_{s_0} - \hat{V}^\pi_{s_0}\\
    \quad&=\quad \sum_{t=0}^{\infty} \gamma^t \langle P^t_{s_0}, R^\pi\rangle - \gamma^t\langle \hat{P}^t_{s_0}, \hat{R}^\pi \rangle\\
    \quad&=\quad \sum_{t=0}^{\infty} \gamma^t\Big(\langle P^t_{s_0}, R^\pi\rangle - \langle \bar{M}^t_{s_0}, R^\pi\rangle + \langle \bar{M}^t_{s_0}, R^\pi\rangle - \langle \bar{M}^t_{s_0}, \hat{R}^\pi\rangle + \langle \bar{M}^t_{s_0}, \hat{R}^\pi\rangle - \langle \hat{P}^t_{s_0}, \hat{R}^\pi \rangle\Big) \\
    \quad&=\quad \sum_{t=0}^{\infty} \gamma^t\langle P^t_{s_0} - \bar{M}^t_{s_0}, R^\pi\rangle + \gamma^t\langle \bar{M}^t_{s_0}, R^\pi - \hat{R}^\pi\rangle + \gamma^t\langle \bar{M}^t_{s_0} - \hat{P}^t_{s_0}, \hat{R}^\pi\rangle.
\end{align*}

By construction, $\bar{M}^t_{s_0}$ is the overlap between $P^t_{s_0}$ and $\hat{P}^t_{s_0}$, and thus and entries of $\bar{M}^t_{s_0} - \hat{P}^t_{s_0}$ are strictly non-positive. Since rewards are likewise non-negative, the third inner product in the above sum is always non-positive. Thus, we can drop this term to significantly tighten our bound.

\begin{align*}
    V^\pi_{s_0} - \hat{V}^\pi_{s_0}\quad&\le\quad \sum_{t=0}^{\infty} \gamma^t\langle P^t_{s_0} - \bar{M}^t_{s_0}, R^\pi\rangle + \gamma^t\langle \bar{M}^t_{s_0}, R^\pi - \hat{R}^\pi\rangle \\
    &\le\quad \sum_{t=0}^{\infty} \gamma^t\lVert P^t_{s_0} - \bar{M}^t_{s_0} \rVert_1 \cdot \lVert R^\pi\rVert_\infty + \gamma^t\lVert \bar{M}^t_{s_0} \rVert_1 \cdot \lVert R^\pi - \hat{R}^\pi\rVert_\infty\\
    &\le\quad \sum_{t=0}^{\infty} \gamma^t\lVert P^t_{s_0} - \bar{M}^t_{s_0} \rVert_1 + \gamma^t\lVert \bar{M}^t_{s_0} \rVert_1 \epsilon_R \\
    &=\quad \sum_{t=0}^{\infty} \gamma^t\lVert P^t_{s_0}\rVert_1 - \gamma^t\lVert\bar{M}^t_{s_0} \rVert_1 + \gamma^t\lVert \bar{M}^t_{s_0} \rVert_1 \epsilon_R \\
    &=\quad \sum_{t=0}^{\infty} \gamma^t + \gamma^t(\epsilon_R - 1)\lVert \bar{M}^t_{s_0} \rVert_1 \\
    &\le\quad \sum_{t=0}^{\infty} \gamma^t + \gamma^t(\epsilon_R - 1)(1 - \epsilon_T/2)^t \\
    &=\quad \frac{1}{1 - \gamma} + (\epsilon_R - 1)\sum_{t=0}^{\infty} (\gamma - \frac{\gamma\epsilon_T}{2})^t \\
    \implies |V^\pi_{s_0} - \hat{V}^\pi_{s_0}| \quad &\le\quad \frac{1}{1 - \gamma} - \frac{1 - \epsilon_R}{1 - \gamma(1 - \epsilon_T/2)}. \quad\quad\quad\quad\quad\quad\quad\quad\quad\quad\quad\quad\quad\qed \\
\end{align*}

This proof makes use of H\"older's inequality to bound inner products with $L_1$ and $L_\infty$ norms, as well as the identity in Equation \ref{eq:p-bar-identity} to split ${\lVert P^t_{s_0} - \bar{M}^t_{s_0}\rVert_1}$ into ${\lVert P^t_{s_0}\rVert_1 - \lVert\bar{M}^t_{s_0}\rVert_1}$. We provide a parallel proof for the finite-horizon undiscounted setting in Appendix \ref{sec:appendix-finite-horizon}. We briefly remark that this bound matches intuition:
\begin{itemize}
    \item When $\gamma=0$, then $|V^\pi_s - \hat{V}^\pi_s| \le \epsilon_R$ since only the first step contributes to value.
    \item When $\epsilon_R = 1$, the MDPs can have completely different reward functions and thus ${|V^\pi_s - \hat{V}^\pi_s| \le \frac{1}{1 - \gamma} = V_{MAX}}$.
    \item When $\epsilon_R = \epsilon_T = 0$, the MDPs are identical and thus ${|V^\pi_s - \hat{V}^\pi_s| = 0}$.
\end{itemize}
\begin{figure*}[t!]
    \begin{subfigure}[b]{0.45\textwidth}
        \centering
        \includegraphics[width=\linewidth]{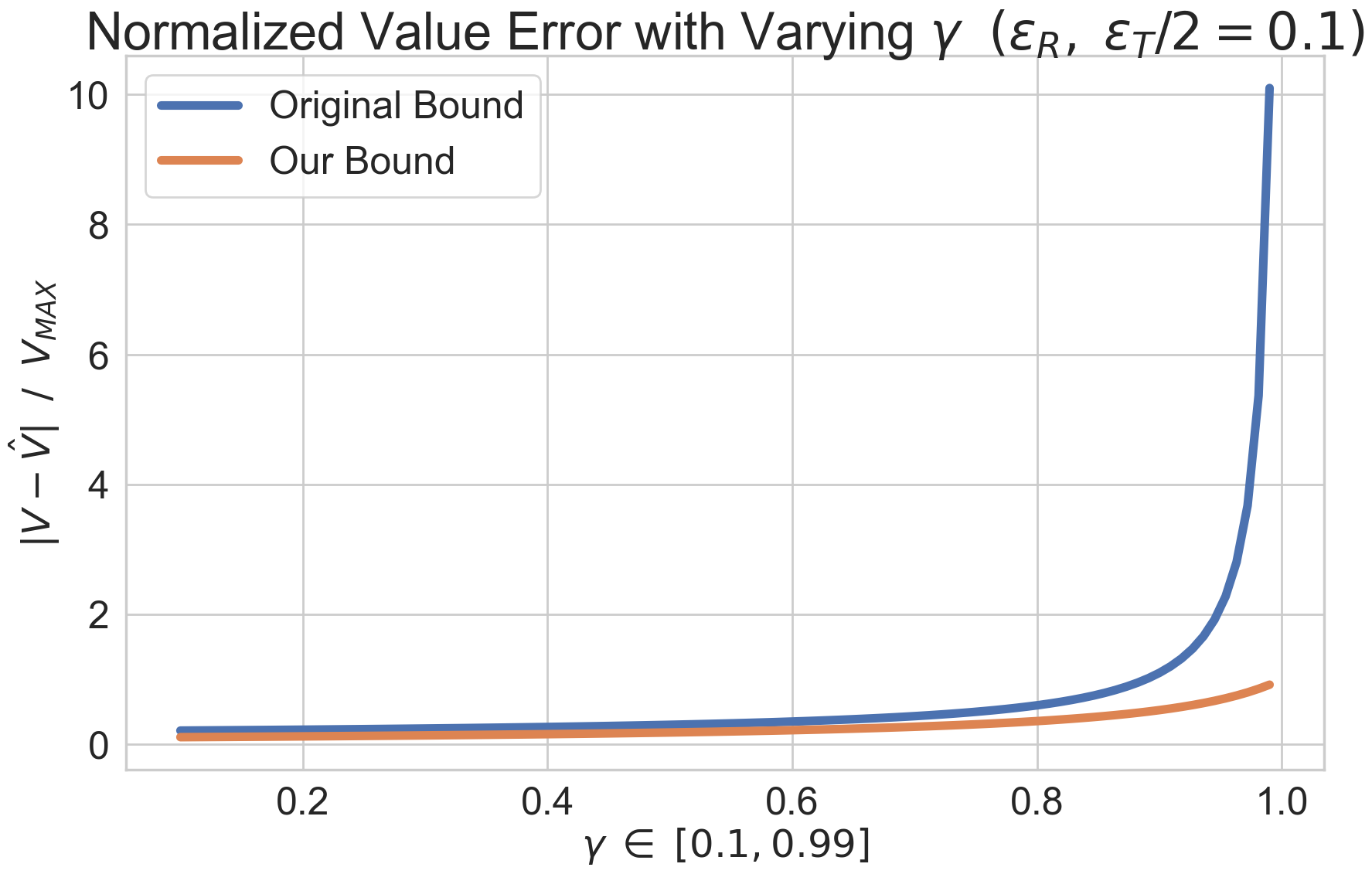}
        \label{fig:value-error-with-varied-gamma}
    \end{subfigure}
    \hfill
    \begin{subfigure}[b]{0.45\textwidth}
        \centering
        \includegraphics[width=\linewidth]{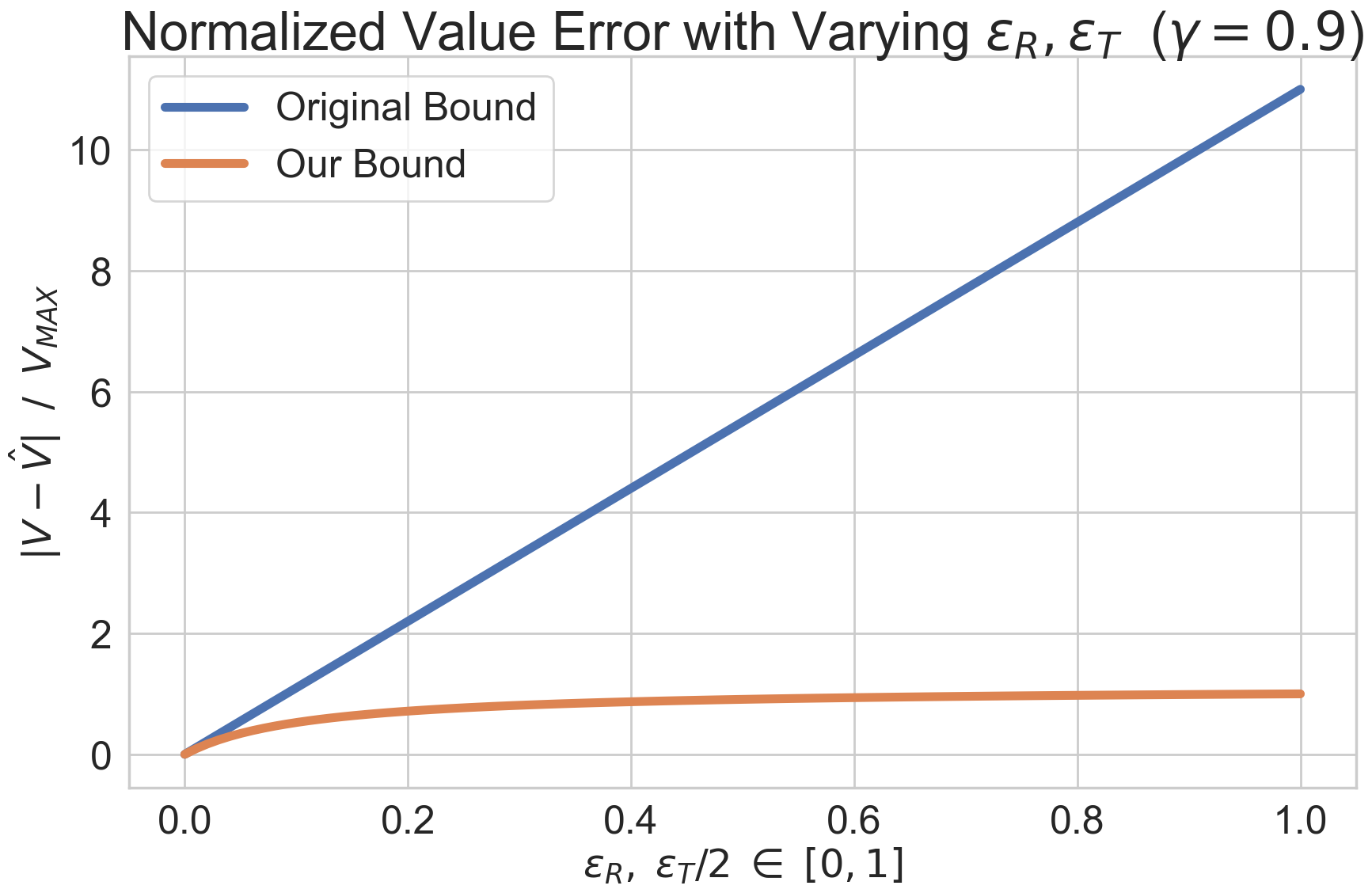}
        \label{fig:value-error-with-varied-epsilon}
    \end{subfigure}
    \vspace{-0.3cm}
    \caption{Bounds on value error given by original simulation lemma as well as our tighter bounds, normalized by $V_{MAX}$. (Left) Bound on value error with increasing gamma shows the original lemma's suboptimality with respect to discount. (Right) Bound on value error with increasing misspecification shows looseness of linear approximation compared to the tight bound.}
    \label{fig:value-error-graphs}
\end{figure*}

Additionally we note the the original simulation lemma can be reproduced as a Taylor expansion of our bound around
$\epsilon_R=0$ and $\epsilon_T=0$,
proving that the original bound is the tightest possible linear approximation to the maximal error as model misspecification approaches 0. Figure \ref{fig:value-error-graphs} presents a comparison of our bound with the original simulation lemma, demonstrating superiority in the large-misspecification and large-discount limits.

\subsection{Proof of Tightness}
\label{subsec:proof-of-tightness}
We now demonstrate that this is the tightest possible bound, including constant factors, by constructing a pair of MDPs with exactly this value error.
$\M$ consists of two states, both of which transition to themselves, with $R(s_1)=1$ and $R(s_2)=0$. We construct $\hat{\M}$ so that $\hat{V}(s_1)$ is as small as possible given $\epsilon_R,\epsilon_T$, by setting $\hat{R}(s_1)=1-\epsilon_R$, and transitioning from $s_1$ to $s_2$ with probability $\epsilon_T/2$ (and thus self-transitions with $\epsilon_T/2$ less probability, so $\lVert P^\pi_{s_1} - \hat{P}^\pi_{s_1}\rVert_1 = \epsilon_T$). Hence, $V(s_0) = \frac{1}{1 - \gamma}$ and $\hat{V}(s_0) = \frac{1 - \epsilon_R}{1 - \gamma(1 - \epsilon_T/2)}$.

Intuitively, this result makes clear the role of $\epsilon_T$ as modifying the discount factor of $\hat{\M}$. A discount can be interpreted as entering an absorbing state with probability $1 - \gamma$ at each timestep \citep{sutton2018reinforcement}.
In $\hat{\M}$, this instead occurs more frequently, with probability $1 - \gamma(1 - \epsilon_T/2)$.

\subsection{Value Loss of Optimal Policy}
The simulation lemma directly applies to bounding the value difference of executing the same policy on two related MDPs. However, in reinforcement learning the task is frequently to learn an \textit{optimal} policy $\pi^*$, that has the following property:
$$\forall \pi, s: V^{\pi^*}_s \ge V^{\pi}_s.$$
It is natural to ask, if one learns the optimal policy $\hat{\pi}^*$ by training on an approximate MDP $\hat{\M}$, how much worse will this policy do than $\pi^*$ when executed on the actual MDP $\M$? In contrast to the simulation lemma, we are comparing the value loss of \textit{different} policies on the \textit{same} MDP. Noting that $\hat{V}^{\hat{\pi}^*}_s \ge \hat{V}^{\pi^*}_s$:
\begin{align*}
    V^{\pi^*}_s - V^{\hat{\pi}^*}_s \quad&=\quad V^{\pi^*}_s + (\hat{V}^{\pi^*}_s - \hat{V}^{\pi^*}_s) + (\hat{V}^{\hat{\pi}^*}_s - \hat{V}^{\hat{\pi}^*}_s) - V^{\hat{\pi}^*}_s\\
    &=\quad (V^{\pi^*}_s - \hat{V}^{\pi^*}_s) + (\hat{V}^{\pi^*}_s - \hat{V}^{\hat{\pi}^*}_s) + (\hat{V}^{\hat{\pi}^*}_s - V^{\hat{\pi}^*}_s) \\
    &\le\quad (V^{\pi^*}_s - \hat{V}^{\pi^*}_s) + 0 + (\hat{V}^{\hat{\pi}^*}_s - V^{\hat{\pi}^*}_s) \\
    &\le\quad |V^{\pi^*}_s - \hat{V}^{\pi^*}_s| + |\hat{V}^{\hat{\pi}^*}_s - V^{\hat{\pi}^*}_s|. \\
\end{align*}

This is simply twice the value error of executing the \textit{same} policy on \textit{different} MDPs. 
Thus, by improving the simulation lemma bound, we similarly tighten the estimated value loss when training on an approximate MDP. Similar results are common in inverse RL, e.g., \citet{burchfiel2016distance}, and have been noted in the context of the simulation lemma as well \citep{jiang2018notes}.

\subsection{Application to Hierarchy}
Analogs to the simulation lemma exist throughout the reinforcement learning literature; here, we present an extension of our proof to one such instance in the field of hierarchical reinforcement learning. We use the formalism of $\phi$-relative options \citep{abel2020value}, a form of approximately value preserving state and action abstractions.

Let $\mathcal{O}_\phi^*$ be a set of options $o^*$ over abstract states $s_\phi\in\mathcal{S}_\phi$, that can be composed to form a policy that is optimal in the base MDP. Let $\hat{\mathcal{O}}_{\phi}$ be a set of options that approximates $\mathcal{O}_{\phi}^*$ in that
\begin{align*}
    &\quad\forall o^* \in \mathcal{O}^*_\phi \;\;\exists \hat{o} \in \hat{\mathcal{O}}_{\phi}:\\
    &{\forall s,s' \;\;\; |P^{o^*}_{s, s'} - P^{\hat{o}}_{s,s'}| \le \epsilon_T}\;\text{ and }\;{|R^{o^*}_{s} - R^{\hat{o}}_{s}| \le \epsilon_R},
\end{align*}

where $R^o_{s}$ and $P^{o}_{s,s'}$ represent the reward and multi-time models of \citet{sutton1999between}. We define $V^{\pi_{o^*}}$ as the value of executing the best policy over $\mathcal{O}^*_\phi$, and $V^{\pi_{\hat{o}}}$ as the value of executing an approximately equivalent policy using options from $\hat{\mathcal{O}}_{\phi}$.
By bounding probability distances we arrive at the following relation:

$$|V^{\pi_{o^*}}_s - V^{\pi_{\hat{o}}}_s| \le \frac{R_{MAX}}{1 - \gamma} - \frac{R_{MAX} - \epsilon_R}{1 - \gamma + (|S|-1)\epsilon_T}.$$

This improves on the existing bound \citep{abel2020value}:

$$|V^{\pi_{o^*}}_s - V^{\pi_{\hat{o}}}_s| \le \frac{\epsilon_R + |S|\epsilon_T R_{MAX}}{(1 - \gamma)^2},$$

in much the same way as our original result improves upon the simulation lemma. A proof, more complete definitions, and an example demonstrating tightness are deferred to Appendix \ref{sec:appendix-hierarchy-bound}. The main difference in applying our technique to this domain is careful treatment of the multi-time transition function, where $\sum_{s'} P^{o}_{s,s'} \ne 1$.

\section{Conclusion}
The simulation lemma is a widely used result in reinforcement learning that quantifies the effect of model misspecification on value. We demonstrate that the originally provided bound is quite loose, becoming vacuous when applied to large discount factors frequently used in reinforcement learning. In this work we present  a version of this lemma that is optimally tight, along with an example application of this method to hierarchical reinforcement learning.
We expect that our bound can be applied to a variety of results throughout the literature, and that the general proof technique can be useful in other domains.

\section*{Acknowledgements}
We would like to thank George Konidaris for valuable input during the early stages of this work, as well as Tuluhan Akbulut and Ruo Yu Tao for helping inspire the question we ask here. This material is based upon work supported by the National Science Foundation Graduate Research Fellowship under grant \#2040433 and ARO grant \#W911NF2210251.

\bibliography{main}
\bibliographystyle{rlc}

\appendix

\section{Full proof of Simulation Lemma}
\label{sec:appendix-full-old-sim-lemma-proof}
For completeness, we include the proof of the simulation lemma found in \citet{jiang2018notes}. We adopt notation from Section \ref{sec:method}.
\begin{align*}
    |V^\pi_s - \hat{V}^\pi_s| \quad&=\quad |R^\pi_s + \gamma \langle P^\pi_s, V^\pi \rangle - \hat{R}^\pi_s - \gamma \langle \hat{P}^\pi_s, \hat{V}^\pi \rangle|  \\
    &\le \quad\epsilon_R + \gamma |  \langle P^\pi_s, V^\pi \rangle - \langle \hat{P}^\pi_s, V^\pi \rangle + \langle \hat{P}^\pi_s, V^\pi \rangle - \langle \hat{P}^\pi_s, \hat{V}^\pi \rangle|  \\
    &= \quad\epsilon_R + \gamma |  \langle P^\pi_s, V^\pi - \frac{\mathbf{1}}{2(1-\gamma)} \rangle - \langle \hat{P}^\pi_s, V^\pi - \frac{\mathbf{1}}{2(1-\gamma)}\rangle + \langle \hat{P}^\pi_s, V^\pi \rangle - \langle \hat{P}^\pi_s, \hat{V}^\pi \rangle|  \\
    &\le \quad\epsilon_R + \gamma \lVert P^\pi_s - \hat{P}^\pi_s \rVert_1 \cdot \lVert V^\pi - \frac{\mathbf{1}}{2(1-\gamma)}\rVert_\infty + \gamma \lVert \hat{P}^\pi_s \rVert_1 \cdot  \lVert V^\pi - \hat{V}^\pi \rVert_\infty \\
    &\le \quad\epsilon_R + \frac{\gamma \epsilon_T}{2(1 - \gamma)} +  \gamma \lVert V^\pi - \hat{V}^\pi \rVert_\infty \\
    \implies |V^\pi_s - \hat{V}^\pi_s| \quad&\le\quad \frac{\epsilon_R}{1 - \gamma} + \frac{\gamma\epsilon_T}{2(1-\gamma)^2}. \\
\end{align*}

This proof makes use of H\"older's inequality to bound inner products with $L_1$ and $L_\infty$ norms, as well as centers the value $0 \le V^\pi_s \le \frac{1}{1-\gamma}$ through subtracting the midpoint for improved bounds.

\section{Application to the Finite-Horizon Setting}
\label{sec:appendix-finite-horizon}
We now extend our improved bound to the finite-horizon, undiscounted setting, where an agent interacts with an environment for $H$ steps.
One difference in this setting is that policies are conditioned on timestep as well as state; hence we define $\pi = [\pi_0,\dots,\pi_{H-1}]$.
Existing bounds in the finite-horizon setting establish a relationship between values at subsequent timesteps. Noting that $0 \le V^{\pi}_{h,s} \le H - h$ (and defining $V^\pi_{H,s} = 0$), Then,
\begin{align*}
|V^{\pi}_{h,s} - \hat{V}^{\pi}_{h,s}| \quad&=\quad |R^{\pi_h}_s + \langle P^{\pi_h}_s, V^{\pi}_{h+1}\rangle - \hat{R}^{\pi_h}_s - \langle\hat{P}^{\pi_h}_s, \hat{V}^{\pi}_{h+1}\rangle| \\
&\le\quad \epsilon_R + |\langle P^{\pi_h}_s, V^{\pi}_{  h+1}\rangle - \langle \hat{P}^{\pi_h}, V^\pi_{h+1} \rangle + \langle \hat{P}^{\pi_h}, V^\pi_{h+1} \rangle - \langle\hat{P}^{\pi_h}_s, \hat{V}^{\pi}_{h+1}\rangle| \\
&=\quad \epsilon_R + |\langle P^{\pi_h}_s, V^{\pi}_{h+1}- \frac{H-h-1}{2}\cdot\mathbf{1}\rangle - \langle \hat{P}^{\pi_h}, V^\pi_{h+1} - \frac{H-h-1}{2}\cdot\mathbf{1}\rangle \\
&\quad\quad\quad\;\,+ \langle \hat{P}^{\pi_h}, V^\pi_{h+1} \rangle - \langle\hat{P}^{\pi_h}_s, \hat{V}^{\pi}_{h+1}\rangle| \\
&\le\quad \epsilon_R + \lVert P^{\pi_h}_s - \hat{P}^{\pi_h}_s\rVert_1 \cdot \lVert V^{\pi}_{h+1} - \frac{H-h-1}{2}\cdot\mathbf{1}\rVert_\infty + \lVert V^{\pi}_{h+1} - \hat{V}^{\pi}_{h+1}\rVert_\infty\\
&\le\quad \epsilon_R + \epsilon_T \frac{H-h-1}{2} + \lVert V^{\pi}_{h+1} - \hat{V}^{\pi}_{h+1}\rVert_\infty\\
\implies |V^{\pi}_{h,s} - \hat{V}^{\pi}_{h,s}| \quad&\le\quad \sum\limits_{i=h}^{H-1} \epsilon_R + \epsilon_T \frac{H-i-1}{2}\\
\implies
|V^{\pi}_{0,s} - \hat{V}^{\pi}_{0,s}| \quad&\le\quad \epsilon_R H + \epsilon_T\frac{H(H-1)}{4}
\end{align*}

For our bound, the only change from the discounted setting is replacing the discounted infinite sums of Section \ref{subsec:main-proof} with finite undiscounted ones. Redefining  $P^t = \prod_{0\le i<t} P^{\pi_i}$, and WLOG assuming that $V^{\pi}_{0,s_0} \ge \hat{V}^{\pi}_{0,s_0}$ we can show:

\begin{align*}
    |V^{\pi}_{0,s_0} - \hat{V}^{\pi}_{0,s_0}| \quad&=\quad V^{\pi}_{0,s_0} - \hat{V}^{\pi}_{0,s_0}\\
    &= \quad\sum_{t=0}^{H-1} \langle P^t_{s_0}, R^{\pi_t}\rangle - \langle \hat{P}^t_{s_0}, \hat{R}^{\pi_t} \rangle \\
    &= \quad\sum_{t=0}^{H-1} \langle P^t_{s_0} - \bar{M}^t_{s_0}, R^{\pi_t}\rangle + \langle \bar{M}^t_{s_0}, R^{\pi_t} - \hat{R}^{\pi_t}\rangle + \langle \bar{M}^t_{s_0} - \hat{P}^t_{s_0}, \hat{R}^{\pi_t}\rangle \\
    &\le \quad\sum_{t=0}^{H-1} \langle P^t_{s_0} - \bar{M}^t_{s_0}, R^{\pi_t}\rangle + \langle \bar{M}^t_{s_0}, R^{\pi_t} - \hat{R}^{\pi_t}\rangle \\
    &\le \quad\sum_{t=0}^{H-1} \lVert P^t_{s_0} - \bar{M}^t_{s_0} \rVert_1 \cdot \lVert R^{\pi_t}\rVert_\infty + \lVert \bar{M}^t_{s_0} \rVert_1 \cdot \lVert R^{\pi_t} - \hat{R}^{\pi_t}\rVert_\infty\\
    &\le \quad\sum_{t=0}^{H-1} \lVert P^t_{s_0} - \bar{M}^t_{s_0} \rVert_1 + \lVert \bar{M}^t_{s_0} \rVert_1 \epsilon_R \\
    &= \quad\sum_{t=0}^{H-1} \lVert P^t_{s_0}\rVert_1 - \lVert\bar{M}^t_{s_0} \rVert_1 + \lVert \bar{M}^t_{s_0} \rVert_1 \epsilon_R \\
    &\le \quad\sum_{t=0}^{H-1} 1 + (\epsilon_R - 1)(1 - \epsilon_T/2)^t \\
    \implies |V^{\pi}_{0,s_0} - \hat{V}^{\pi}_{0,s_0}| \quad&\le\quad H - (1 - \epsilon_R)\frac{2}{\epsilon_T}(1 - (1 - \epsilon_T/2)^H) \\
\end{align*}

Again, we note that Taylor expanding this relation at $\epsilon_R=0$ and $\epsilon_T=0$ recovers the original bound.

\section{Proof of Hierarchy Bound}
\label{sec:appendix-hierarchy-bound}
This proof exactly mirrors the one in the main body, with additional care taken to handle multi-time models. We first describe the $\phi$-relative options framework (definitions largely taken from \citet{abel2020value}), and then provide a tighter bound on value loss.

An \textit{option} $o \in \mathcal{O}$ is an abstract action defined by the tuple ($\mathcal{I}_o, \beta_o, \pi_o$), where $\mathcal{I}_o \subseteq \S$ is the subset of the state space the option can initiate in, $\beta_0 \subseteq \S$ is the subset the option terminates in, and $\pi_o$ is a policy.
For a given state abstraction $\phi: \S \rightarrow \S_\phi$, an option $o_\phi$ is said to be $\phi$-relative if and only if ${\exists s_\phi \in \S_\phi}$ such that 
$$s \in s_\phi \implies s \in \mathcal{I}_{o_\phi}\quad\quad s \notin s_\phi \implies s \in \mathcal{\beta}_{o_\phi} \quad\quad \forall s \in s_\phi,\;\;\pi_{o_\phi} (s) \rightarrow \Delta{(\A)}$$
In words, a $\phi$-relative option is one that executes from anywhere in one abstract state, and terminates upon leaving that abstract state. Furthermore, $\mathcal{O}_\phi$ denotes a set of only $\phi$-relative options, with at least one option that executes at each abstract state.

Let $\mathcal{O}_{\phi}^*$ be a set of $\phi$-relative options $o^*$ that can be composed to form an optimal policy in the base MDP. Let $\hat{\mathcal{O}}_{\phi}$ be a set of options that approximates $\mathcal{O}_{\phi}^*$ in that

\begin{equation}
\begin{aligned}
    &\quad\forall o^* \in \mathcal{O}^*_\phi \;\;\exists \hat{o} \in \hat{\mathcal{O}}_{\phi}:\\
    &{\forall s,s'\;\;\;|P^{o^*}_{s, s'} - P^{\hat{o}}_{s,s'}| \le \epsilon_T}\;\text{ and }\;{|R^{o^*}_{s} - R^{\hat{o}}_{s}| \le \epsilon_R}
    \label{eq:p-and-r-hierarchy-appendix}
\end{aligned}
\end{equation}

where $R^o_{s}$ and $P^o_{s,s'}$ represent the multi-time reward and transition functions described in \citet{sutton1999between}:
$$R^o_{s} = \mathbb{E}_{a \sim o}[\sum_{t=0}^\infty \gamma^t r_{t}]\quad\quad P^o_{s,s'} = \sum_{t=1}^\infty \gamma^t \Pr(s_t=s', t_\beta = t).$$

In words, $R^o_{s}$ is the expected discounted reward accumulated over the course of an option execution, and $P^o_{s,s'}$ is the total discounted probability that an option terminates in $s'$ when starting from $s$. Crucially, $\sum_{s' \in \S} P^o_{s,s'} \le \gamma < 1$. We also note that the $\epsilon_T$ bound is per-entry, not per-vector. This was the form of the conditions in the original simulation lemma \citep{kearns2002near}, which was replaced with a vectorized version in subsequent work \citep{kakade2003exploration}.

Since ${\lVert P^{o}_s \rVert_1}$ may take on different values for different options and starting states, we can no longer directly use a relation similar to Equation \ref{eq:tvd-l1-equivalence}. However, we can augment the MDP by adding an absorbing state $s_x$, and modify each option such that
$$R^o_{s_x} = 0\quad,\quad P^o_{s,s_x} = \gamma - \sum_{s' \neq s_x} P^o_{s,s'}.$$

By doing this, ${\lVert P^{o}_s \rVert_1 = \gamma}$ without modifying the behavior of the given option in the base MDP. This allows our proof to proceed treating options in roughly the same way as we do actions in the main body. Noting that since  $P^o_{s,s}\equiv 0$ by construction, for two options $o^*, \hat{o}^*$ satisfying the relations of Equation \ref{eq:p-and-r-hierarchy-appendix} we have that:

$$| P^{o^*}_{s,s_x} - P^{\hat{o}}_{s,s_x}| \le \sum_{s'\neq s_x,s}|P^{o^*}_{s,s'} - P^{\hat{o}}_{s,s'}| \le (|S|-1)\epsilon_T.$$

Thus we can recover a condition similar to that of Equation \ref{eq:original-sim-lemma-t-condition}:

$$\lVert P^{o^*}_s - P^{\hat{o}}_s\rVert_1 = \sum_{s' \in \S+s_x}|P^{o^*}_{s,s'} - P^{\hat{o}}_{s,s'}| \le 2(|S| - 1)\epsilon_T.$$

Due to the addition of $s_x$, we can now describe the above bound in terms of overlap. Defining $\bar{P}^{o^*,\hat{o}}_{s,s'} = \min(P^{o^*}_{s,s'}, P^{\hat{o}}_{s,s'})$, we can produce a similar relation to Equation \ref{eq:tvd-sim-lemma-conditions}:

$$\lVert \bar{P}^{o^*,\hat{o}}_s\rVert_1 \ge \gamma - (|S|-1)\epsilon_T.$$

Let $\Pi_{\mathcal{O}_\phi}$ be the set of abstract policies representable by $\mathcal{O}_\phi$.
Let $\pi_{o^*}$ be a policy within $\Pi_{\mathcal{O}^*_\phi}$ that is optimal in the base MDP. Let $\pi_{\hat{o}}$ be a policy in $ \Pi_{\hat{\mathcal{O}}_\phi}$ produced by replacing each $o^*$ chosen by $\pi_{o^*}$ with an option $\hat{o}$ satisfying the relations of Equation \ref{eq:p-and-r-hierarchy-appendix}. Then, we can follow the same algebraic steps as in the main body to produce the following bound:

$$|V^{\pi_{o^*}}_s - V^{\pi_{\hat{o}}}_s| \le \frac{R_{MAX}}{1 - \gamma} - \frac{R_{MAX} - \epsilon_R}{1 - \gamma + (|S|-1)\epsilon_T}.$$

\subsection{Proof of Tightness}
We can generate an abstract MDP that achieves this bound using a similar recipe as in Section \ref{subsec:proof-of-tightness}. We construct an abstract MDP where each option $o^*$ transitions uniformly to each other state with discounted probability $\frac{\gamma}{|S|-1}$, receiving a reward of $R_{MAX}$. We then construct a new set of options that uniformly transition with discounted probability $\frac{\gamma}{|S|-1} - \epsilon_T$, receiving reward $R_{MAX} - \epsilon_R$. This exactly reproduces the provided bound.

\end{document}